\documentclass[sigconf]{acmart}
\usepackage{multirow}

\AtBeginDocument{%
  }

\copyrightyear{2025}
\acmYear{2025}
\setcopyright{cc}
\setcctype{by-sa}
\acmConference[MM '25]{Proceedings of the 33rd ACM International Conference on Multimedia}{October 27--31, 2025}{Dublin, Ireland}
\acmBooktitle{Proceedings of the 33rd ACM International Conference on Multimedia (MM '25), October 27--31, 2025, Dublin, Ireland}\acmDOI{10.1145/3746027.3755335}
\acmISBN{979-8-4007-2035-2/2025/10}
 
\begin{document}

\title{TASR: Timestep-Aware Diffusion Model for Image Super-Resolution}

\author{Qinwei Lin}
\authornote{Co-First Authors.}
\email{lqw22@mails.tsinghua.edu.cn}
\affiliation{%
  \institution{Shenzhen International Graduate School, Tsinghua University}
  \city{Shenzhen}
  \country{China}
}

\author{Xiaopeng Sun}
\authornotemark[1]
\email{xpsun@stu.xidian.edu.cn}
\affiliation{%
  \institution{Meituan Inc.}
  \city{Shenzhen}
  \country{China}
}

\author{Yu Gao}
\authornotemark[1]
\email{nkugaoyu@163.com}
\affiliation{%
  \institution{Meituan Inc.}
  \city{Shenzhen}
  \country{China}}

\author{Yujie Zhong}
\affiliation{%
  \institution{Meituan Inc.}
  \city{Beijing}
  \country{China}
}

\author{Zheng Zhao}
\affiliation{%
 \institution{Meituan Inc.}
 \city{Beijing}
 \country{China}}

\author{Dengjie Li}
\affiliation{%
  \institution{Meituan Inc.}
 \city{Beijing}
 \country{China}}

\author{Haoqian Wang}
\authornote{Corresponding Author.}
\email{wanghaoqian@tsinghua.edu.cn}
\affiliation{%
  \institution{Shenzhen International Graduate School, Tsinghua University}
  \city{Shenzhen}
  \country{China}}


\renewcommand{\shortauthors}{Qinwei Lin et al.}

\begin{abstract}
Diffusion models have recently achieved outstanding results in the field of image super-resolution. 
These methods typically inject low-resolution (LR) images via ControlNet. 
In this paper, we first explore the temporal dynamics of information infusion through ControlNet, revealing that the input from LR images predominantly influences the initial stages of the denoising process. 
Leveraging this insight, we introduce a novel timestep-aware diffusion model that adaptively integrates features from both ControlNet and the pre-trained Stable Diffusion (SD). 
Our method enhances the transmission of LR information in the early stages of diffusion to guarantee image fidelity and stimulates the generation ability of the SD model itself more in the later stages to enhance the detail of generated images. 
To train this method, we propose a timestep-aware training strategy that adopts distinct losses at varying timesteps and acts on disparate modules. 
Experiments on benchmark datasets demonstrate the effectiveness of our method.
\end{abstract}

\ccsdesc[500]{Computing methodologies~Image processing}

\keywords{Diffusion Model, Image Super-Resolution}



\maketitle
\section{Introduction}
\label{sec:intro}
Image super-resolution (ISR) aims to reconstruct high-resolution (HR) images from their low-resolution (LR) counterparts.
The effectiveness of methods based on generative adversarial networks (GANs) has been demonstrated in previous works~\cite{wang2021realesrgan, zhang2021bsrgan, liang2021swinir,wang2018esrgan,zhang2019ranksrgan}. 
However, when dealing with severely degraded LR images, the generated HR images contain numerous visual artifacts and lack realistic details, resulting in low visual quality.

Recently, denoising diffusion probabilistic models (DDPMs)~\cite{ho2020ddpm,dhariwal2021diffusion,song2020ddim,peebles2023scalable_dit,chen2023pixart,song2020scoresde,saharia2022photorealisticimagen,nichol2021glide,rombach2022highldm,esser2024scalingsd3} have achieved remarkable performance in the field of image generation, gradually replacing GANs~\cite{goodfellow2020generativegans} in a series of downstream image generation tasks.
Therefore, some works~\cite{wu2024seesr, chen2024cassr, lin2024diffbir, yu2024supir, yang2023pasd, qi2023tip, sun2024coser, wang2024stableSR, yue2024resshift,liu2022blind,xie2024addsr} have leveraged large-scale pre-trained diffusion models to solve the ISR tasks. 
These models are mainly based on the ControlNet~\cite{zhang2023controlnet}, which is used to inject the LR image as a condition into the latent feature space of a pre-trained diffusion model. 
This category of diffusion-based methods typically requires sampling over multiple timesteps during the generation process.
Previous works~\cite{choi2022perception, hu2024ella, agarwal2023matte, liu2024faster, sun2023improving} have pointed out that diffusion models primarily generate low-frequency semantic content in the initial stages of the denoising process, while gradually generating high-frequency details in the later stages.
However, it remains unclear whether ControlNet exhibits similar patterns when applied to diffusion-based ISR.

\begin{figure}
  \centering
  \includegraphics[width=\linewidth]{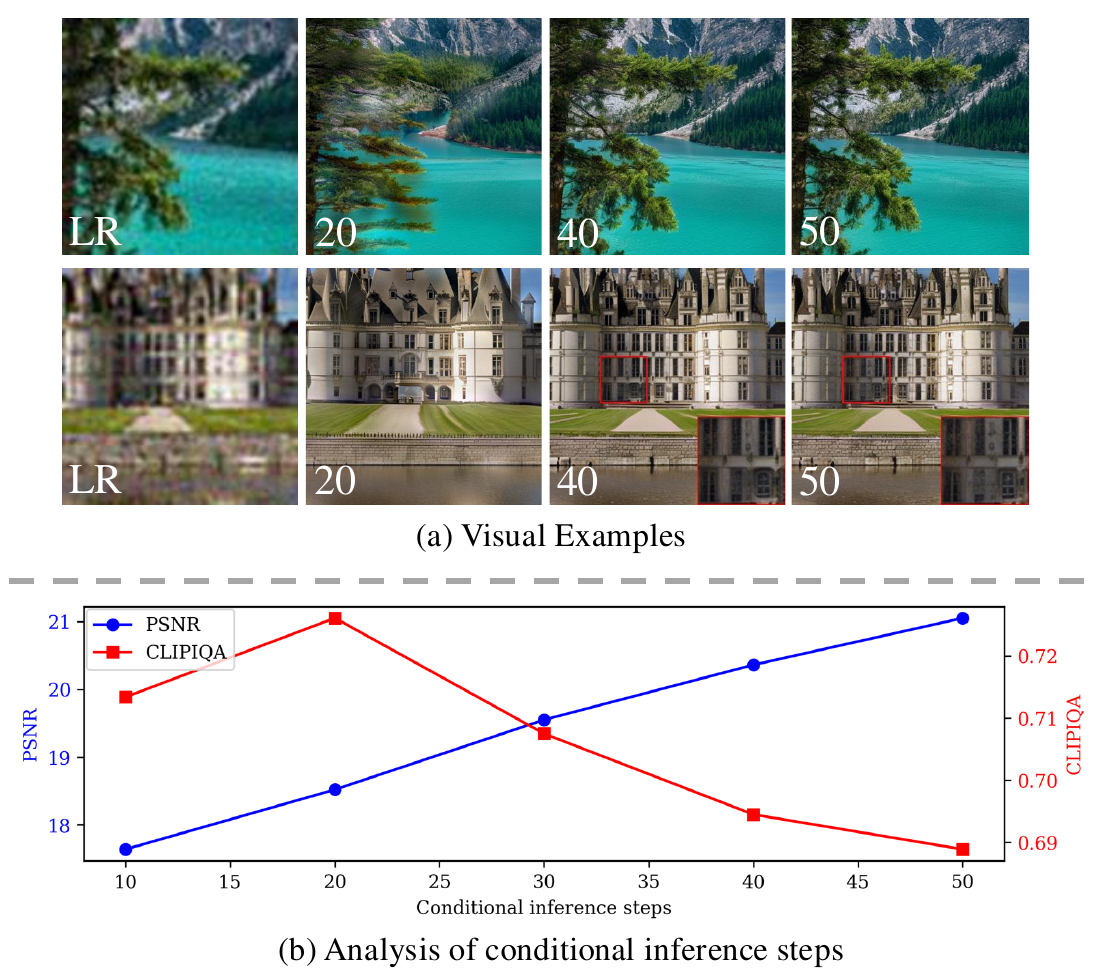}
  \caption{Effectiveness of ControlNet at different timesteps in the denoising process. (a) Generated HR images at different conditional inference steps. ``LR" denotes the given LR images, ``20" indicates that the ControlNet features are introduced only in the first 20 steps of the inference process. (b) Analysis of the generated HR images under different conditional inference steps based on PSNR and CLIPIQA~\cite{wang2022exploringclipiqa}.}
  \label{fig:example_of_timestep}
\end{figure}

To this end, we conduct a simple experiment based on DiffBIR~\cite{lin2024diffbir} to explore the effectiveness of ControlNet at different timesteps in the denoising process.
As shown in Fig.~\ref{fig:example_of_timestep}, as the conditional steps of ControlNet increase, the fidelity of the generated HR images correspondingly improves, as indicated by a gradual increase in PSNR.
On the other hand, the visual quality may deteriorate, as indicated by a decrease in the CLIPIQA~\cite{wang2022exploringclipiqa,radford2021learningclip} scores.
As shown in the visual examples, introducing ControlNet during the initial inference timesteps significantly improves the structural consistency between the generated HR image and the given LR image.
Interestingly, disabling ControlNet during the late stages of inference (e.g., the last 10 timesteps) has minimal impact on the visual fidelity of the generated outputs. 
In some cases, the image details (e.g., windows on buildings) are even better.
The above experiments indicate that ControlNet primarily affects the structural information of the generated HR images in the early stages of the denoising process. 
However, in the later stages of the denoising process, this constraining effect diminishes and may potentially impede the generation of intricate details.

The above observations inspire us to design a timestep-aware adapter that adaptively integrates ControlNet features with diffusion features at different timesteps.
Based on the role of different timesteps, we anticipate that the adapter will emphasize the structural and color information of the image by increasing the weights of ControlNet features in the early stages of denoising.
Meanwhile, in the later stages, the adapter is expected to enhance the generation of fine-grained image details by focusing more on the diffusion model features.
To achieve this goal, we propose a timestep-aware training strategy to optimize our method, applying L1 loss functions from early timesteps to ensure image fidelity and introducing the CLIPIQA score in the later stages to enhance visual quality.
Furthermore, by recognizing the characteristics corresponding to each loss function, different loss functions are used to optimize the corresponding modules within the model separately. 
Overall, the main contributions are summarized as follows:
\begin{itemize}
    \item [$\bullet$] We propose a novel diffusion-based method for image super-resolution, which designs an adapter to dynamically control the feature fusion process between the ControlNet features and the diffusion features. This control is guided by the timesteps in the denoising process, allowing for a more nuanced integration of information.
    \item [$\bullet$] We introduce a timestep-aware training strategy that employs distinct loss functions to optimize the ControlNet and Adapter modules at different timesteps.
    \item [$\bullet$] Experiments on benchmark datasets demonstrate the effectiveness and superiority of our method.
\end{itemize}

\section{Related Work}
\label{sec:related}

\subsection{Diffusion-based Image Super-Resolution}
Diffusion models have shown excellent performance in the field of image generation. 
Therefore, recent research works~\cite{lin2024diffbir,wu2024seesr,yang2023pasd,wang2024stableSR,yu2024supir,yue2024resshift,sun2024coser,qi2023tip,xie2024addsr} utilized powerful pre-trained text-to-image diffusion models~\cite{rombach2022stablediffusion, dhariwal2021diffusion, podell2023sdxl}  as generative priors to tackle image super-resolution tasks.
DiffBIR~\cite{lin2024diffbir} initially proposed a restoration module to remove degradation noise from the LR images, then utilized a pre-trained SD~\cite{rombach2022stablediffusion} as a generative module. 
The denoised LR images are used as control signals for ControlNet~\cite{zhang2023controlnet} to generate the final HR image.
SeeSR~\cite{wu2024seesr} and PASD~\cite{yang2023pasd} both proposed a degradation-aware prompt extractor to extract semantic prompts from LR images and use these prompts as auxiliary conditional information to guide the denoising process together with the LR images.
SUPIR~\cite{yu2024supir} collected a large-scale dataset as training data and used a more powerful pre-trained text-to-image diffusion model, SDXL~\cite{podell2023sdxl}, as a generative prior.  
It also leveraged a multimodal large language model (MLLM)~\cite{liu2024llava} to extract textual descriptions to guide the generation of HR images.
From the perspective of model architecture, most of these works are based on ControlNet~\cite{zhang2023controlnet}, which receives LR images as conditions and passes ControlNet features into the pre-trained diffusion model to guide the generation of corresponding HR images.
Whereas, these methods do not take into account the role of ControlNet in the generation of HR images at different timesteps.
Therefore, how to enhance the visual quality of generated images from a temporal perspective based on the ControlNet architecture is the main objective of this paper.

\subsection{Temporal Analysis of Diffusion Model}
During inference, diffusion models start from random noise and generate corresponding images through multiple steps of sampling and denoising.
Recently, some studies~\cite{balaji2022ediff,choi2022perception,hu2024ella,agarwal2023matte,liu2024faster,sun2023improving,sun2024rfsr} have found that at different timesteps, diffusion models focus on different aspects during the denoising process: in the early stages of denoising, model primarily generates low-frequency information, such as the semantics and structure of the image, while in the later stages, model tends to generate high-frequency information, such as the edges and details of the image.
ELLA~\cite{hu2024ella} proposed a timestep-aware semantic connector that dynamically extracted control information of different frequencies from LLM text features at various denoising stages.
T-GATE~\cite{liu2024faster} investigated the role of attention in the denoising process from a temporal perspective and improved computational efficiency by caching and reusing attention operations at different timesteps during inference.
MATTE~\cite{agarwal2023matte} decomposed multiple attributes (e.g., color, object, layout, style) of the reference image from both the temporal dimension and the network layer dimension, injecting layout and color attributes at different timesteps during the denoising process to achieve attribute-guided image synthesis.
Inspired by these works, we analyze the role of ControlNet from the temporal dimension and propose a timestep-aware adapter between ControlNet and the pre-trained diffusion model to fuse ControlNet features with diffusion features adaptively.

\begin{figure*}
  \centering
  \includegraphics[width=\linewidth]{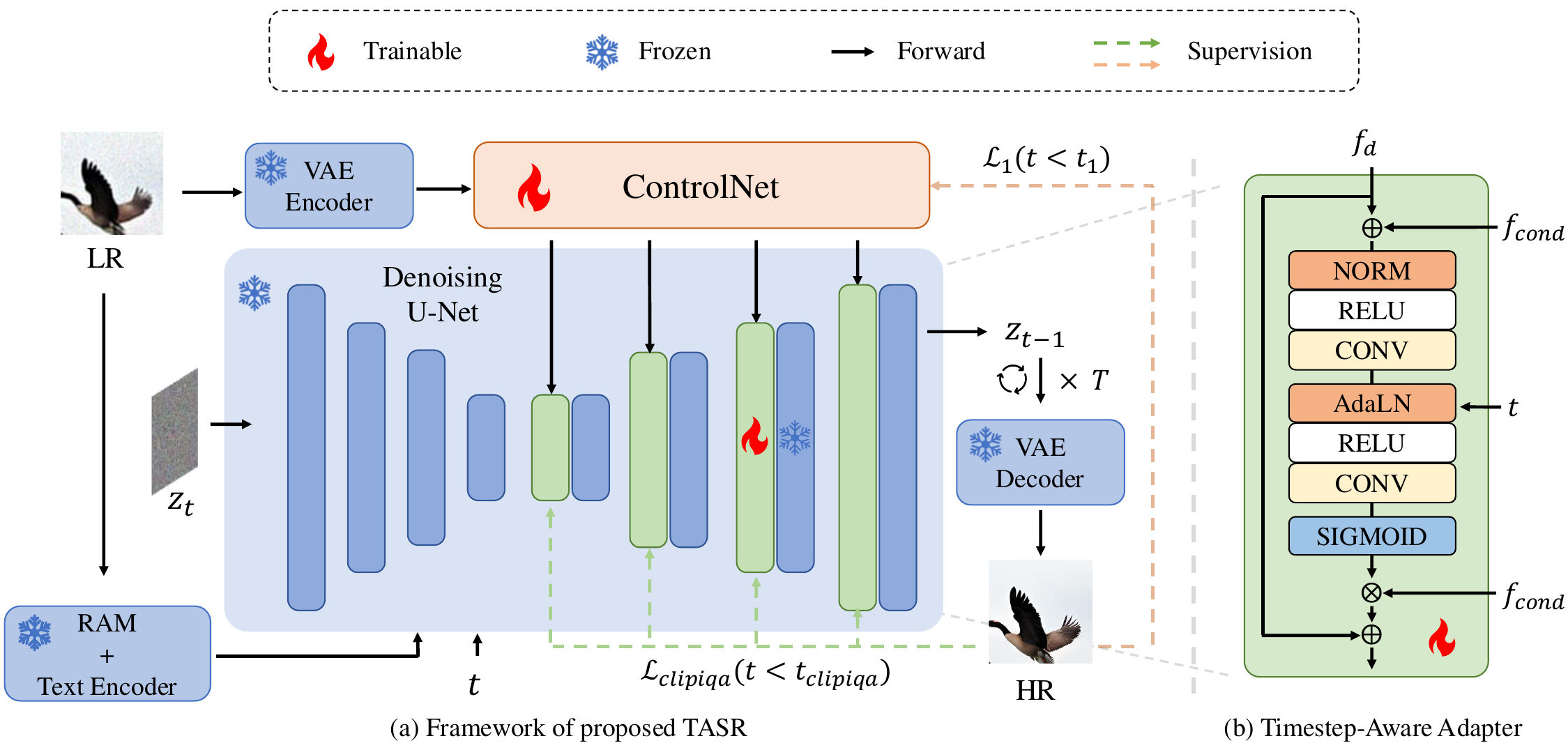}
  \caption{Overview of TASR. (a) TASR is built on the ControlNet and introduces the Timestep-Aware Adapter in each decoder block of the denoising U-Net. The $\mathcal{L}_1$ is used to optimize the ControlNet when $t < t_{1} $, while the $\mathcal{L}_{clipiqa}$ is applied to optimize the proposed Timestep-Aware Adapter when $t < t_{clipiqa}$. (b) Architecture details of the Timestep-Aware Adapter.}
  \label{fig:overview}
\end{figure*}
\section{Method}
\label{sec:method}

\subsection{Overview}
In this work, we propose a timestep-aware SR (TASR) method aimed at improving the quality of generated HR images by employing adaptive feature fusion between ControlNet and the diffusion model.
As shown in Fig.~\ref{fig:overview}, our proposed TASR mainly consists of a pre-trained SD model, corresponding ControlNet, and a timestep-aware adapter.
The parameters of the pre-trained SD model are frozen during the entire training stage. 
The VAE~\cite{rombach2022stablediffusion} encoded LR image features are used as the condition input for ControlNet. 
Meanwhile, we use the RAM~\cite{zhang2024ram} to extract text prompts from the given LR images and obtain the text features encoded by the frozen CLIP text encoder~\cite{rombach2022stablediffusion}.
These CLIP text features are fed into the model as additional semantic prompts.
In addition, the designed timestep-aware Adapter is inserted between ControlNet and the pre-trained SD UNet decoder. 
The specific design of each module is detailed in Sec.~\ref{sec:tasr}.
The entire training process is divided into two stages to ensure the effectiveness and stability of ControlNet and the adapter during training.
In the first stage, we optimize only the ControlNet parameters using the SR training dataset, thereby ensuring the effectiveness of ControlNet features.
In the second stage, based on the patterns observed in the denoising process, we design a timestep-aware training strategy to optimize ControlNet and the adapter separately, as described in Sec.~\ref{sec:training}.

\subsection{TASR}\label{sec:tasr}
\noindent \textbf{ControlNet.}
Similar to previous work~\cite{lin2024diffbir, yang2023pasd,wu2024seesr}, we utilize the powerful pre-trained large-scale text-to-image SD model as the image generation module in our model and employ the ControlNet to inject the VAE-encoded LR image feature as an additional condition into the decoder of SD to generate the corresponding HR image.
Following~\cite{zhang2023controlnet}, ControlNet is constructed by creating trainable copies of the pre-trained U-Net encoder and middle blocks, and injecting the conditional ControlNet features into the pre-trained U-Net decoder blocks via a zero convolution layer.
To fully leverage the guidance of text prompts on image generation in SD, we utilize a pre-trained image tagging model~\cite{zhang2024ram} to extract semantic prompt information from LR images.
These prompts are encoded into text features $\boldsymbol{c}$ with the frozen CLIP text encoder and injected into the SD model to guide the denoising process.

\noindent \textbf{Timestep-Aware Adapter}
Based on the empirical observations in Fig.~\ref{fig:example_of_timestep}, injecting ControlNet conditional features during the early stages of the denoising process, i.e., semantic-planning phase~\cite{liu2024faster}, can improve the structural consistency of the generated HR images. 
However, in the later stages of denoising, i.e., the fidelity-improving phase~\cite{liu2024faster}, the role of ControlNet features weakens and may even negatively impact the generation of HR image details.
These observations inspire us to evaluate the role of ControlNet at different timesteps and how to inject ControlNet features into SD in a timestep-aware manner. 
To this end, we design a timestep-aware adapter that predicts a control weight map based on the current timestep to achieve the dynamic feature fusion of ControlNet and SD.
The specific structure of the timestep-aware adapter is illustrated in Fig.~\ref{fig:overview} (right). 
Each adapter consists of two stacked convolutional layers with ReLU layers and normalization layers.
The timesteps are injected through the AdaLN~\cite{hu2024ella,peebles2023scalable_dit,perez2018film} layer, and finally, the adapter outputs the control weight map via a sigmoid function.
The adapter takes the U-Net feature $\boldsymbol{f_d}$ from the previous layer, the skip connection feature $\boldsymbol{f}_{cond}$ from ControlNet and the timestep $t$ as inputs to predict the corresponding control weight $\boldsymbol{\alpha}$ for dynamically injecting the ControlNet feature by $\boldsymbol{f_d}+\boldsymbol{f}_{cond}*\boldsymbol{\alpha}$.

\begin{figure}[t]
  \centering
   \includegraphics[width=\linewidth]{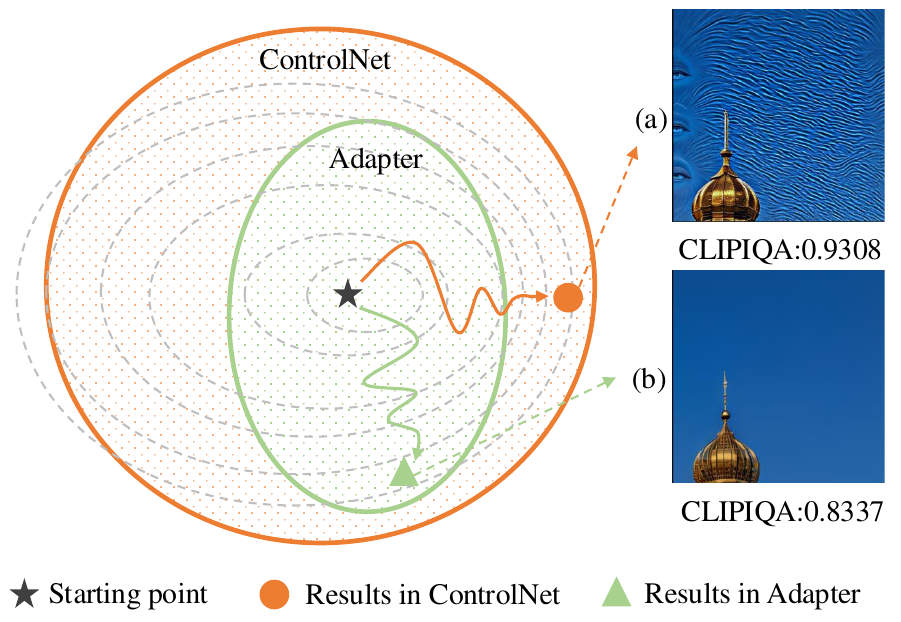}
   \caption{Optimization Space of ControlNet and Adapter.}
   \label{fig:visual_space}
\end{figure}

\subsection{Optimization Objective}\label{sec:training}
\noindent \textbf{Stage I.} 
To ensure the control effectiveness of the ControlNet and improve training stability, we train only the ControlNet in the first stage.
During training, the HR image $\mathbf{I}_{HR}$ and the LR image $\mathbf{I}_{LR}$ are encoded by pre-trained VAE encoder into latent representations $\boldsymbol{z}_{0}$ and $\boldsymbol{z}_{lr}$, respectively.
In addition, we leverage a tag model~\cite{wu2024seesr} to extract text prompts from the LR images and the pre-trained CLIP text encoder to obtain the text features $\boldsymbol{c}$.
The diffusion process~\cite{ho2020ddpm} generates the noisy latent $\boldsymbol{z}_{t}$ by adding Gaussian noise with variance $\beta_t \in (0,1)$ at timestep $t$ to $\boldsymbol{z}_{0}$:
\begin{equation}
    \boldsymbol{z}_{t} = \sqrt{\bar{\alpha}_t} \boldsymbol{z}_0 + \sqrt{1-\bar{\alpha}_t}\epsilon, \quad \epsilon \sim \mathcal{N}(\boldsymbol{0},\mathbf{I}),
\end{equation}
where $\epsilon$ represents a noise map sampled from a normal Gaussian distribution, $\alpha_t=1-\beta_t$, and $\bar{\alpha}_t=\prod^t_{s=1}\alpha_s$.

The ControlNet is initialized as a trainable copy of the pre-trained UNet encoder and middle block, and $\boldsymbol{z}_{t}$ and $\boldsymbol{z}_{lr}$ are concatenated together as the input to the ControlNet.
Given timestep $\boldsymbol{t}$, LR latent $\boldsymbol{z}_{l}$, noisy latent $\boldsymbol{z}_{t}$, text features $\boldsymbol{c}$, we trained our model $\epsilon_\theta$ using denoising loss $\mathcal{L}_{d}$ to predict the noise added to 
$\boldsymbol{z}_t$, as follows:
\begin{equation}
    \mathcal{L}_{d} = \mathbb{E}_{\boldsymbol{z}_{l}, \boldsymbol{z}_{t}, \boldsymbol{c}, \boldsymbol{t},\epsilon \sim \mathcal{N}(0, 1)} \left[ \left\| \epsilon - \epsilon_\theta(\boldsymbol{z}_t, \boldsymbol{t}, \boldsymbol{c}, \boldsymbol{z}_l) \right\| \right].
\end{equation}

\noindent \textbf{Stage II.} 
After the first training stage, ControlNet can learn the correct conditional information from the LR image. 
However, in the second stage, an obvious question arises: how to properly train the adapter to weigh the information from ControlNet in a timestep-aware manner based on the pattern of the diffusion model in the denoising process.
A naive approach is to optimize the adapter using the same training data and the denoising loss as in the previous stage.
However, since the weights of ControlNet have been trained in the first stage, the adapter will infinitely approach an identity function under the same training data and denoising loss function.

As observed in Fig.~\ref{fig:example_of_timestep}, during the early stages of denoising, the model tends to learn image structures and other information from the control information, i.e., $\boldsymbol{z}_{lr}$, while in the later stages of denoising, it focuses on generating high-frequency image details.
Therefore, we propose a timestep-aware training strategy that introduces different loss functions based on the contribution of different stages of the denoising process to guide the image generation process. 
Specifically, we introduce L1 Loss $\mathcal{L}_{1}$ to supervise ControlNet from the early denoising stages (i.e., $ 0 \leq \boldsymbol{t} \leq 800$):
\begin{align}
    \mathcal{L}_{1} &= \left\|\mathbf{I}_{HR}-\mathbf{\hat{I}}\right\|, \\
 \text{where} \quad  \mathbf{\hat{I}} &= \mathbb{D}\left(\frac{x_t - \sqrt{1-\alpha_t}\epsilon_\theta(\boldsymbol{z}_t,\boldsymbol{t},\boldsymbol{c},\boldsymbol{z}_l)}{\sqrt{\alpha}_t}\right),
\end{align}
$\mathbb{D}$ denotes the pre-trained VAE decoder.

In the later stages of denoising (i.e., $ 0 \leq \boldsymbol{t} \leq 200$), we use the non-reference metric CLIP-IQA~\cite{wang2022exploringclipiqa} to evaluate the visual quality of the generated HR images $\mathbf{\hat{I}}$ and use its results as a perception reward to encourage the model to improve the image quality of the generated results.
The specific detailed reward loss is as follows:
\begin{equation}\label{eq:ch3_eq_clipiqa}
    \mathcal{L}_{clipiqa} = 1 - \mathbb{R}(\mathbf{\hat{I}}),
\end{equation}
where $\mathbb{R}$ denotes the CLIP-IQA model.

Without introducing the adapter, directly using the $\mathcal{L}{1}$ and $\mathcal{L}_{clipiqa}$ to optimize the ControlNet is another option. 
We conduct experiments on this scheme, and the experimental results show that the generated images tend to exhibit specific styles, as shown in Fig.~\ref{fig:visual_space} (a). 
We found that this is due to the reward hacking~\cite{skalse2022defining} caused by $\mathcal{L}_{clipiqa}$, resulting in a high CLIPIQA score but poor visual quality. 
As shown in Fig.~\ref{fig:visual_space}, ControlNet has a larger optimization space compared to the proposed adapter, whose optimization space is constrained by the sigmoid function. 
When directly using CLIPIQA as the perceptual reward to train ControlNet, the model can easily fall into the local optimum that aligns with the preference of CLIPIQA. 
The timestep-aware adapter guides image generation by predicting the control weight map $\boldsymbol{\alpha}$ of ControlNet features, with the $\boldsymbol{\alpha}$ values constrained between 0 and 1. 
If only the adapter is optimized, its optimization space is smaller, and the reward hacking point mentioned above falls outside this space.
Therefore, we only utilize $\mathcal{L}_{clipiqa}$ to optimize the parameters of the adapter, thus avoiding perception reward traps.
However, compared to the perception reward loss $\mathcal{L}_{clipiqa}$, the L1 Loss $\mathcal{L}_{1}$ is applied to measure the absolute error between images in a pixel-wise manner, representing the structural differences between images (e.g., color distribution). 
Thus, using L1 Loss $\mathcal{L}_{1}$ to optimize the ControlNet with large parameter space can achieve better fitting results.
To this end, we apply the L1 Loss $\mathcal{L}_{1}$ to optimize ControlNet parameters, while utilizing the perception reward $\mathcal{L}_{clipiqa}$ to optimize the timestep-aware adapter parameters.
Additionally, to enhance the stability of the training process, we adopt an alternating training approach inspired by GANs~\cite{zhang2021bsrgan,wang2021realesrgan}, optimizing the parameters of ControlNet and the adapter in alternate iteration steps to prevent interference between the two modules and allow each to learn more effectively. 
Specifically, we fix the parameters of one module while updating the other, alternately training ControlNet and the adapter. 
The final loss function is as follows:
\begin{equation}
\mathcal{L} =
\begin{cases} 
    \mathcal{L}_d, & \text{if } 
    \begin{aligned} 
        t &\in [t_{1}, 1000] 
    \end{aligned} \\[0.5em]
    \mathcal{L}_d + \lambda_1\mathcal{L}_1, & \text{if } 
    \begin{aligned} 
        t &\in [t_{clipiqa}, t_{1}] 
    \end{aligned} \\[0.5em]
    \mathcal{L}_d + \lambda_1\mathcal{L}_1 + \lambda_{clipiqa}\mathcal{L}_{clipiqa} , & \text{if } 
    \begin{aligned} 
        t &\in [0, t_{clipiqa}] 
    \end{aligned}
\end{cases}
\label{eq:piecewise_function}
\end{equation}
where $\lambda_1$ and $\lambda_{clipiqa}$ are hyperparameters, both set to 0.01. 
The values of $t_1$ and $t_{clipiqa}$ are set to 800 and 200, respectively. 
\section{Experiments}
\label{sec:exp}

\begin{table*}[!ht]
    \centering
    \caption{Quantitative comparison on synthetic and real-world benchmark datasets. \textcolor{red}{\textbf{Red}} and \textcolor{blue}{\underline{blue}} represent the best and the second-best performance, respectively. $\downarrow$ represents the smaller values are better, while $\uparrow$ represents the larger values are better. }
    \label{tab:quantitative_comparsion}
    \renewcommand\arraystretch{1.3}
    \begin{tabular}{c|c|ccc|ccccc|c}
    \toprule
    \multirow{2}{*}{Datasets} &  \multirow{2}{*}{Metrics}  &\multicolumn{3}{c|}{GAN-based methods }
    &\multicolumn{6}{c}{Diffusion-based methods} \\
    && RealESRGAN & BSRGAN & SwinIR
    & SeeSR & PASD   &ResShift  & DiffBIR & SUPIR   
    & Ours \\ \midrule
    \multirow{6}{*}{DIV2K-val} 
    &$\text{PSNR}\uparrow$ 
    &22.34 &\textcolor{blue}{\underline{22.84}} &\textcolor{red}{\textbf{23.29}}
    &21.53 &21.46 &22.02 &21.24 &20.61 
    &20.92 \\
    &$\text{SSIM}\uparrow$ 
    &0.5768 &\textcolor{blue}{\underline{0.5969}} &\textcolor{red}{\textbf{0.6083}} 
    &0.5356 &0.5321 &0.5485 &0.5168 &0.4717 
    &0.5174 \\
    &$\text{LPIPS}\downarrow$ 
    &\textcolor{blue}{\underline{0.3370}} &0.4851 &0.4951
    &\textcolor{red}{\textbf{0.3311}} &0.4408 &0.3644 &0.3691 &0.4203 
    &0.3762\\
    &$\text{MANIQA}\uparrow$ 
    &0.3892 &0.2500 &0.2306 
    &\textcolor{blue}{\underline{0.5094}} &0.3558 &0.3540 &0.4573 &0.4861 
    &\textcolor{red}{\textbf{0.6007}} \\
    &$\text{MUSIQ}\uparrow$ 
    &57.50 &37.71 &31.32 
    &\textcolor{blue}{\underline{67.62}} &52.63 &55.56 &67.43 &54.97 
    &\textcolor{red}{\textbf{68.14}} \\
    &$\text{CLIPIQA}\uparrow$ 
    &0.5365 &0.2869 &0.3110 
    &0.6987 &0.4917 &0.5479 &\textcolor{blue}{\underline{0.7110}} &0.6374 
    &\textcolor{red}{\textbf{0.7681}} \\   \midrule
    
    \multirow{6}{*}{RealSR} 
    &$\text{PSNR}\uparrow$ 
    &25.69 &\textcolor{blue}{\underline{27.27}} &\textcolor{red}{\textbf{27.42}}
    &25.18 &26.56 &26.42 &25.22 &23.74 
    &23.79 \\
    &$\text{SSIM}\uparrow$ 
    &0.7618 &\textcolor{blue}{\underline{0.7983}} &\textcolor{red}{\textbf{0.7999}}
    &0.7200 &0.7614 &0.7569 &0.7028 &0.6631 
    &0.6650 \\
    &$\text{LPIPS}\downarrow$ 
    &\textcolor{red}{\textbf{0.2172}} &0.2312 &0.2440 
    &0.2354 &\textcolor{blue}{\underline{0.2217}} &0.2385 &0.2577 &0.2871 
    &0.2986 \\
    &$\text{MANIQA}\uparrow$ 
    &0.3743 &0.3269 &0.2816 
    &\textcolor{blue}{\underline{0.5427}} &0.3875 &0.3969 &0.4583 &0.5025 
    &\textcolor{red}{\textbf{0.6113}} \\
    &$\text{MUSIQ}\uparrow$ 
    &60.17 &53.12 &45.22 
    &\textcolor{blue}{\underline{69.78}} &59.10 &60.18 &66.62 &61.56 
    &\textcolor{red}{\textbf{69.94}} \\
    &$\text{CLIPIQA}\uparrow$ 
    &0.4444 &0.2952 &0.3166 
    &0.6611 &0.4829 &0.5563 &\textcolor{blue}{\underline{0.6700}} &0.6565
    &\textcolor{red}{\textbf{0.7076}} \\ \midrule
    
    \multirow{6}{*}{DRealSR} 
    &$\text{PSNR}\uparrow$ 
    &28.64 &\textcolor{blue}{\underline{29.91}} &\textcolor{red}{\textbf{30.36}}
    &28.17 &29.05 &28.78 &26.87 &25.00
    &27.25 \\
    &$\text{SSIM}\uparrow$ 
    &0.8052 &\textcolor{blue}{\underline{0.8394}} &\textcolor{red}{\textbf{0.8496}}
    &0.7674 &0.793 &0.7878 &0.7116 &0.6416 
    &0.7381 \\
    &$\text{LPIPS}\downarrow$ 
    &\textcolor{red}{\textbf{0.2121}} &0.2569 &0.2571 
    &\textcolor{blue}{\underline{0.2346}} &0.2371 &0.2508 &0.3045 &0.3340
    &0.2869 \\
    &$\text{MANIQA}\uparrow$ 
    &0.3448 &0.2771 &0.2621 
    &\textcolor{blue}{\underline{0.5146}} &0.3753 &0.3517 &0.4548 &0.4980 
    &\textcolor{red}{\textbf{0.5551}} \\
    &$\text{MUSIQ}\uparrow$ 
    &54.17 &41.76 &36.48 
    &\textcolor{blue}{\underline{64.93}} &52.60 &52.49 &63.34 &59.66 
    &\textcolor{red}{\textbf{65.23}} \\
    &$\text{CLIPIQA}\uparrow$ 
    &0.4418 &0.3145 &0.3460 
    &\textcolor{blue}{\underline{0.6810}} &0.5035 &0.5429 &0.6680 &0.6791 
    &\textcolor{red}{\textbf{0.7155}} \\ \midrule

    \multirow{3}{*}{RealLR200} 
    &$\text{MANIQA}\uparrow$ 
    &0.3677 &0.2968 &0.2152 
    &\textcolor{blue}{\underline{0.5048}} &0.4324 &0.4170 &0.4511 &0.4673 
    &\textcolor{red}{\textbf{0.6150}} \\
    &$\text{MUSIQ}\uparrow$ 
    &62.94 &52.19 &35.81
    &\textcolor{blue}{\underline{69.50}} &66.73 &62.17 &66.02 &64.86 
    &\textcolor{red}{\textbf{71.05}} \\
    &$\text{CLIPIQA}\uparrow$ 
    &0.5392 &0.3674 &0.3682 
    &\textcolor{blue}{\underline{0.6811}} &0.6575 &0.6392 &0.6760 &0.6156 
    &\textcolor{red}{\textbf{0.7660}} \\
    \bottomrule
    \end{tabular}
\end{table*}

\subsection{Dataset and Evaluation Metric}
\noindent \textbf{Datasets.}
Following ~\cite{wu2024seesr,yang2023pasd,lin2024diffbir}, we train our method on DIV2K~\cite{agustsson2017ntire_div2k}, DIV8K~\cite{gu2019div8k}, Flickr2K~\cite{timofte2017ntire_flickr2k}, OST~\cite{wang2018recovering_ost}, and the first 10,000 face images from FFHQ~\cite{karras2019style_ffhq}, and use the degradation pipeline of RealESR-GAN~\cite{wang2021realesrgan} to generate HR-LR image pairs for training.
We evaluate the performance of our method on both synthetic and real-world datasets. 
The synthetic dataset is generated from the DIV2K validation set, where we use the same degradation pipeline in the training process to randomly crop 3K image patches to $128\times128$ as LR images.
For real-world test datasets, we evaluate the DrealSR~\cite{wei2020drealsr}, RealSR~\cite{wei2020drealsr}, and RealLR200~\cite{wu2024seesr}datasets. 
In both DrealSR and RealSR datasets, each image is center-cropped to obtain LR images of $128\times128$ resolution. 
The resolution of each HR image in both the training and test sets is $512\times512$.
The RealLR200 dataset is built by SeeSR, which comprises 200 LR images from recent works and a series of real-world scenes.

\noindent \textbf{Metrics.}
We perform a comprehensive and effective quantitative evaluation of ISR methods using a series of widely used reference and non-reference metrics. 
Among the reference-based metrics, PSNR and SSIM~\cite{wang2004imagessim} (calculated on the Y channel in the YCbCr space) are fidelity metrics, while LPIPS~\cite{zhang2018unreasonablelpips} are quality assessment metrics. 
MANIQA~\cite{yang2022maniqa_musiq}, MUSIQ~\cite{yang2022maniqa_musiq}, and CLIPIQA~\cite{wang2022exploringclipiqa} are non-reference image quality assessment (IQA) metrics.

\begin{figure}[!htb]
  \centering
  \includegraphics[width=\linewidth]{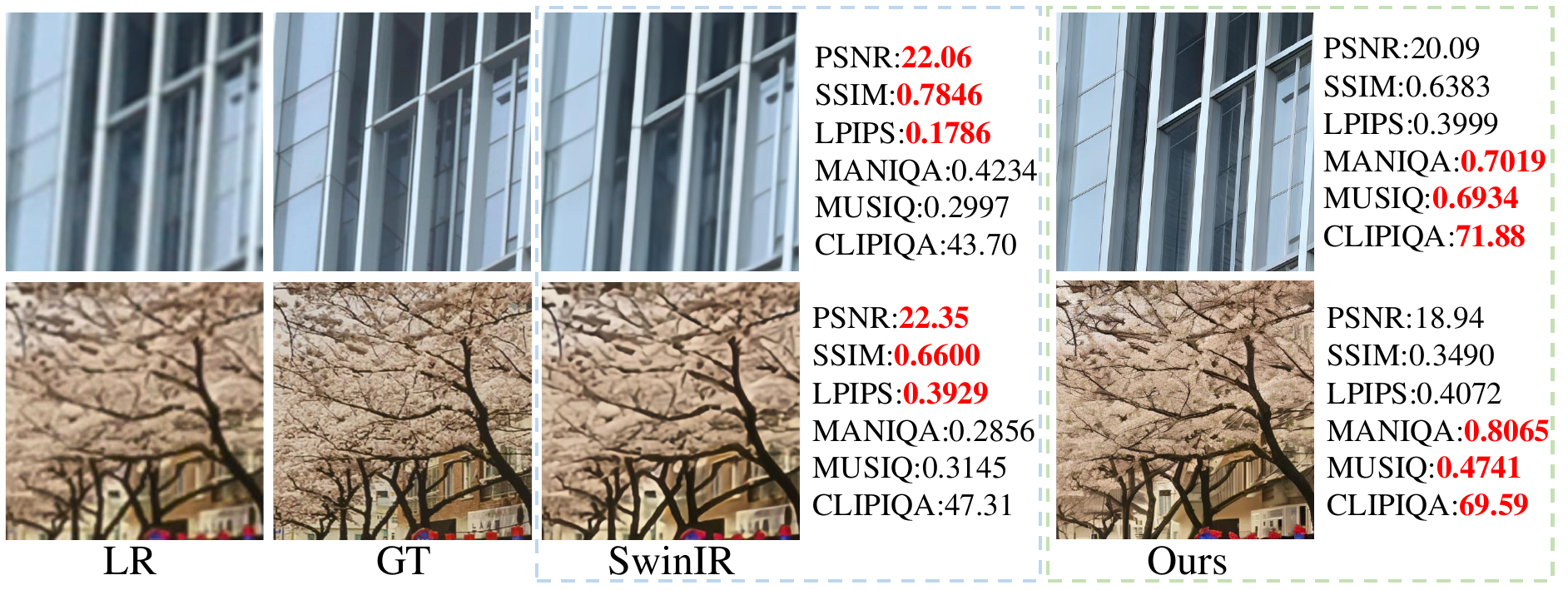}
  \caption{\textbf{Drawbacks of reference metrics.}}
  \label{fig:drawback_metrics}
\end{figure}

\begin{figure*}[!t]
  \centering
  \includegraphics[width=\linewidth]{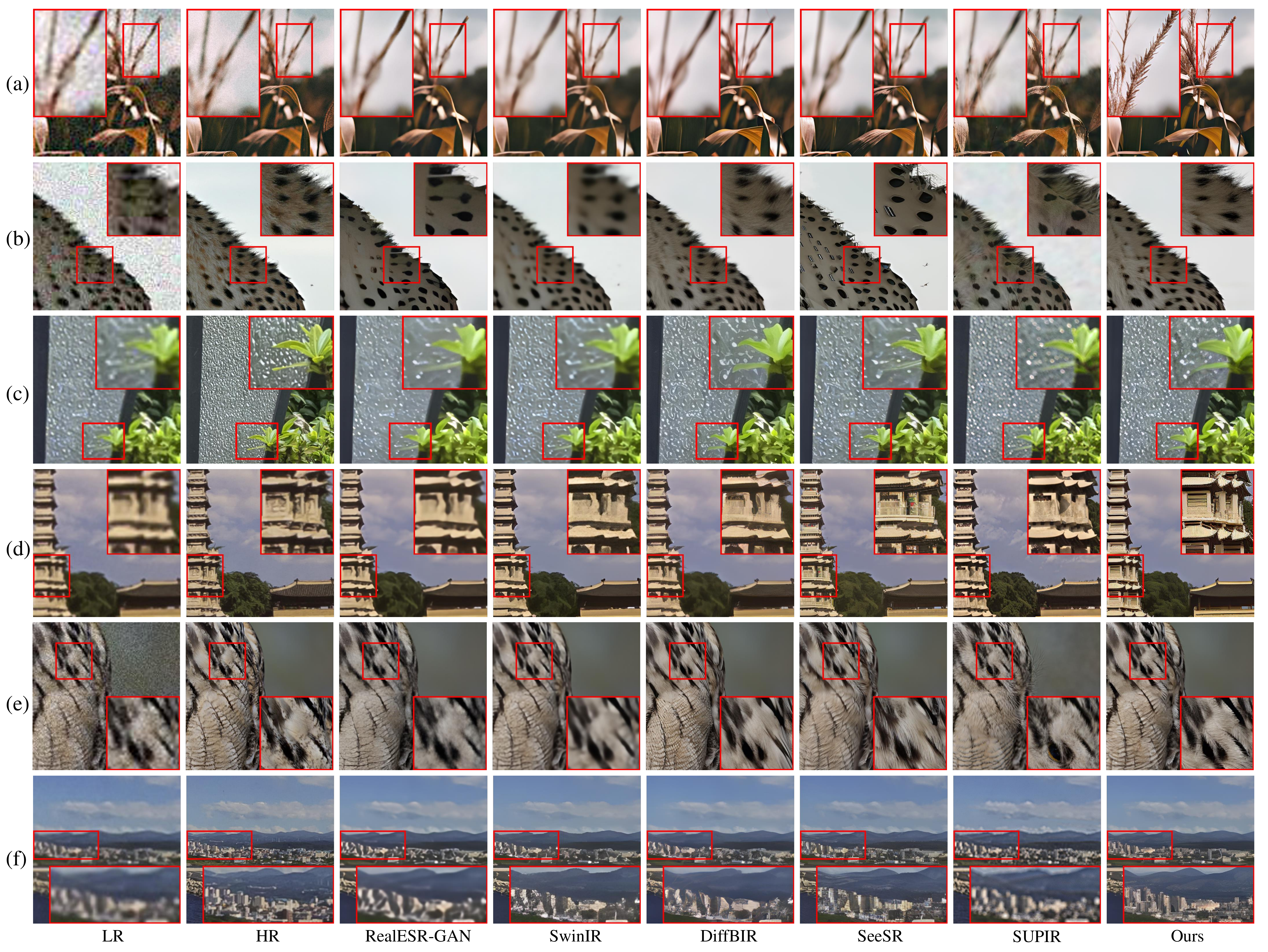}
  \caption{Qualitative comparisons with different ISR methods on both synthetic and real-world test datasets.}
  \label{fig:qual_cmp}
\end{figure*}

\subsection{Implementation Details}
We use the pre-trained SD-v2.1~\cite{rombach2022stablediffusion} model as the base SD model, with the ControlNet module being initialized using the base SD model parameters. 
In the first stage of training, we fine-tune the ControlNet module for 20K iterations, and in the second stage, we fine-tune both the ControlNet and Adapter for 100K iterations with the same training dataset. 
During the two-stage training, we use the AdamW~\cite{kinga2015methodadam} optimizer with a weight decay of 1e-2, a batch size of 32, and a learning rate of 1e-5. 
All experiments are conducted at a resolution of 512×512 on 8 NVIDIA A100 GPUs.
During inference, we employ a classifier-free guidance strategy, generating higher-quality images through given negative prompts without additional training. 
The classifier-free guidance scale is set to 4.5, and we use a spaced DDPM sampling schedule~\cite{nichol2021improvedddim} with 20 time steps.

\subsection{Comparison with State-of-the-art Methods}
To validate the effectiveness of our method, we compared it with several state-of-the-art GAN-based and diffusion-based ISR methods, namely RealESRGAN~\cite{wang2021realesrgan}, BSRGAN~\cite{zhang2021bsrgan}, SwinIR~\cite{liang2021swinir}, SeeSR~\cite{wu2024seesr}, PASD~\cite{yang2023pasd}, ResShift~\cite{yue2024resshift}, DiffBIR~\cite{lin2024diffbir}, and SUPIR~\cite{yu2024supir}.

\noindent \textbf{Quantitative Comparisons.}
Tab.~\ref{tab:quantitative_comparsion} presents the quantitative comparison on both synthetic and real-world test datasets.
As observed, our method significantly outperforms other state-of-the-art (SOTA) methods on non-reference metrics such as MANIQA, MUSIQ, and CLIPIQA across all datasets, generating higher-quality HR images. 
For example, on the DRealSR dataset, TASR outperforms the second-best method, SeeSR by 7.8\%, 0.4\%, and 5.0\% on MANIQA, MUSIQ, and CLIPIQA metrics, respectively.
On the RealLR200 dataset, compared to the second-best method, TASR achieves improvements of 21.83\% and 12.46\% on the MANIQA and CLIPIQA metrics, respectively.
Furthermore, GAN-based methods surpass diffusion-based methods on reference metrics such as PSNR and SSIM. 
We attribute this phenomenon to the fact that diffusion-based methods leverage powerful generative priors to generate details that are more perceptually realistic to humans,  but this comes at the expense of fidelity to the LR images.
As noted in previous works~\cite{wu2024seesr,yang2023pasd}, this fundamental trade-off results in high perception quality but low scores on reference metrics.
Fig.~\ref{fig:drawback_metrics} demonstrates our method can generate high-quality images aligned with human perception while lagging in reference metrics in some scenes.

\begin{figure}[!t]
  \centering
  \includegraphics[width=\linewidth]{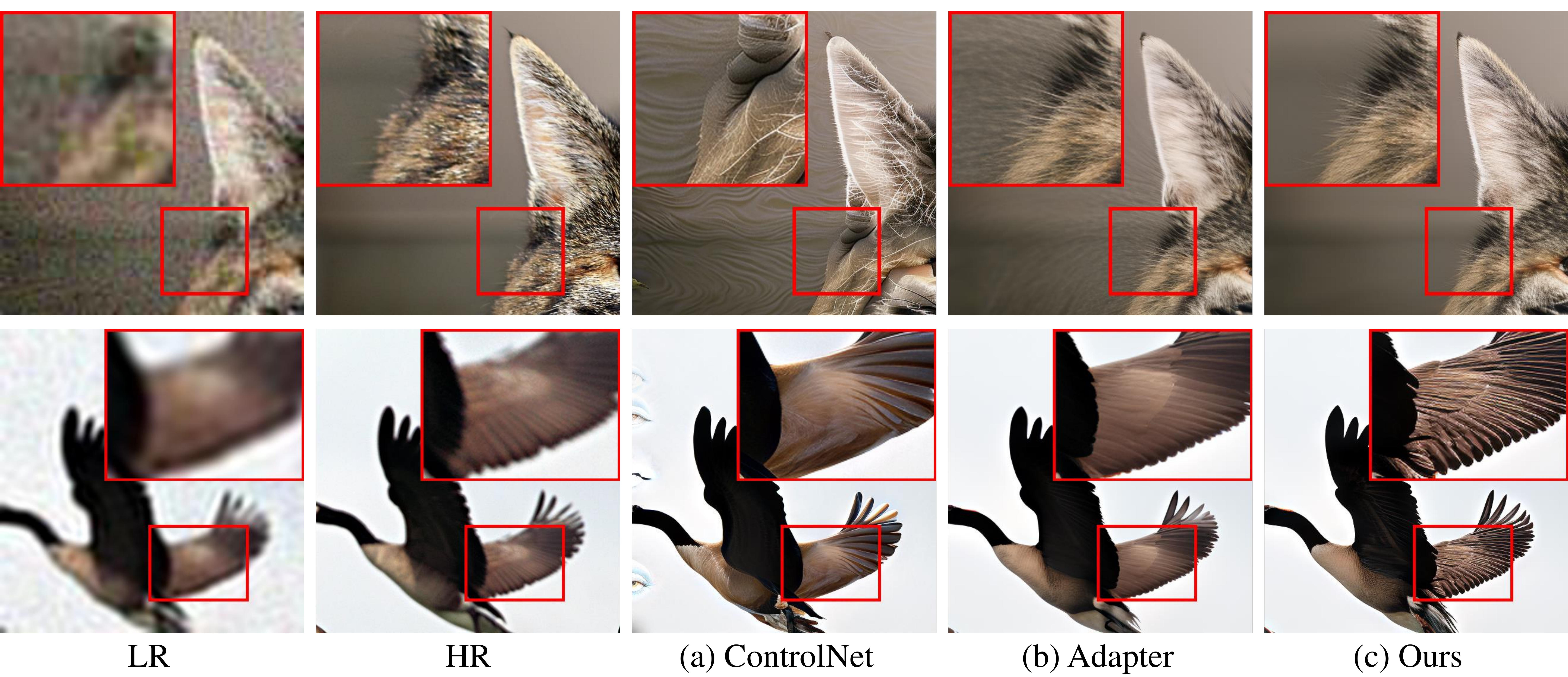}
  \caption{Visual comparison for Timestep-Aware Adapter.}
  \label{fig:ablation_adapter}
\end{figure}

\noindent \textbf{Qualitative Comparisons.}
In Fig.~\ref{fig:qual_cmp}, we present some visual comparison results on both synthetic and real-world test datasets.
As observed, GAN-based approaches such as Real-ESRGAN and SwinIR fail to generate fine image details compared to diffusion-based methods, and the resulting HR images still exhibit a certain degree of degradation.
Meanwhile, our method outperforms other diffusion-based methods in terms of image structural fidelity and detail richness. 
As shown in Fig.~\ref{fig:qual_cmp} (a,d,f), the HR images generated by other diffusion-based methods still contain a certain degree of blurring and artifacts in the complex region.
In contrast, our method effectively removes these degradations and generates more refined image details,  such as the realistic ear of wheat, the edge details of the building, and clearer urban landscapes.
Furthermore, compared to other methods, our approach generates image details with more accurate semantics.
As shown in Fig.~\ref{fig:qual_cmp} (b), RealESR-GAN, SeeSR, and SUPIR all fail to generate accurate animal hair textures.
In Fig.~\ref{fig:qual_cmp} (e), SUPIR mistakenly generates the black spots on the feathers as eyes.
In contrast, employing the proposed timestep-aware training strategy, our method can generate HR images with richer details while maintaining structural consistency with the LR images.


\begin{figure}[!t]
  \centering
  \includegraphics[width=\linewidth]{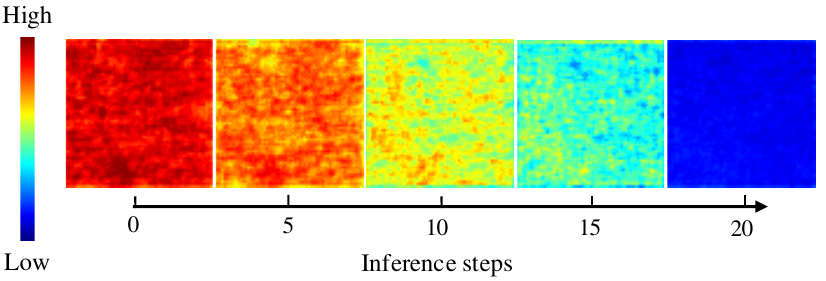}
  \caption{Visual examples of control weight map. The visual control weight map is obtained by averaging the control weight maps from all scenes in the DIV2K-val test dataset. }
  \label{fig:ablation_control_weight_map}
\end{figure}

\begin{figure*}[!htbp]
  \centering
  \includegraphics[width=\linewidth]{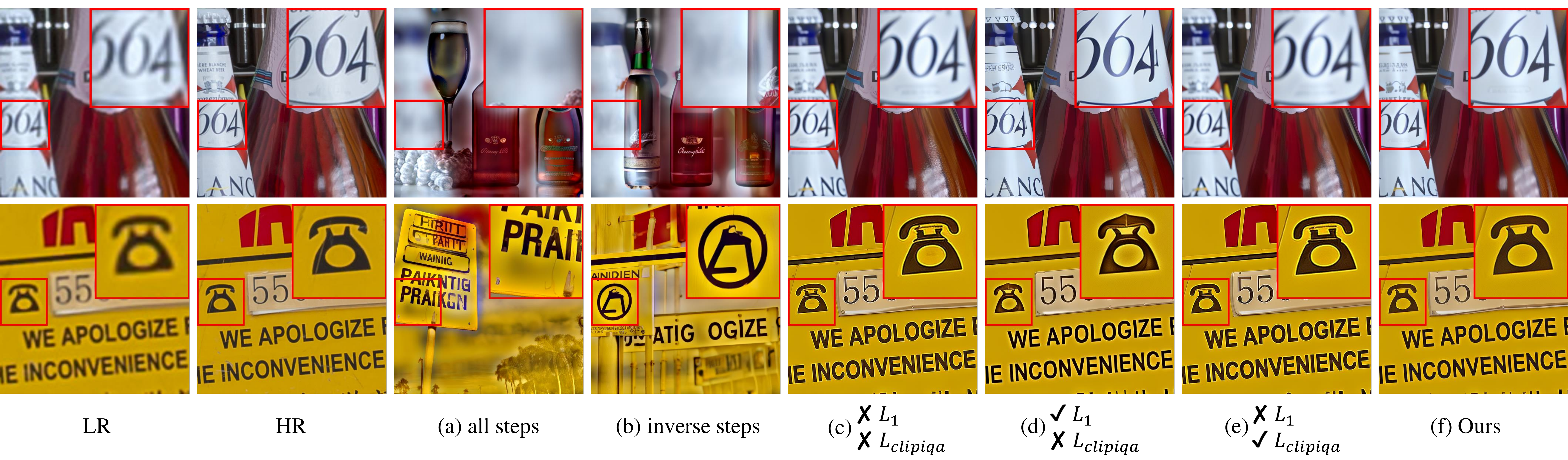}
  \caption{Visual comparison for ablation studies on Loss Functions and Timestep-Aware training strategy.}
  \label{fig:ablation_steps}
\end{figure*}

\subsection{Ablation Study}
To validate the effectiveness of our proposed method, we conduct experiments on the DIV2K-val test dataset.

\noindent \textbf{Model Architecture.}
We employ various architectures to validate the effectiveness of our proposed model structure for the adapter.
Firstly, we removed all time-adaptive normalization layers from the adapter and directly utilized the U-Net features $\boldsymbol{f}_d$ and the skip connection features $\boldsymbol{f}_{cond}$ to predict the control weight map $\boldsymbol{\alpha}$.
This modification is denoted as `w/o timestep'.
As shown in Tab.~\ref{tab:ablation_adapter}, by incorporating timesteps into the adapter, our method achieves better performance in all the metrics compared to the variant without timesteps.
In addition, we replace our adapter structure with a transformer-based architecture similar to ELLA~\cite{hu2024ella}, denoted as `Transformer'.
The variant with the Transformer block has declined in the non-reference metrics. 

\begin{table}[!ht]
    \centering
    \caption{Ablations of Model Architecture.}
    \label{tab:ablation_adapter}
    \begin{tabular}{c cccc} \toprule
      &$\text{LPIPS}\downarrow$ 
      &$\text{MANIQA}\uparrow$ 
      &$\text{MUSIQ}\uparrow$ 
      &$\text{CLIPIQA}\uparrow$ 
      
  \\ \midrule 
    w/o timestep 
    &0.3965  
    &0.5856 &67.38   &0.7668  \\
    Transformer
    &0.3619  
    &0.5790 &67.64   &0.7583 \\
    ControlNet  
    &0.4632
    &0.6876 &74.90   &0.9345  \\
    Adapter  
    &0.3561  
    &0.5628 &66.75   &0.7501  \\
    Ours            
    &0.3762  
    &0.6007 &68.14   &0.7681 \\ \bottomrule
    \end{tabular}
\end{table}

\noindent \textbf{Training for Different Module.}
Firstly, we remove the timestep-aware adapter and optimize only the parameters of ControlNet during the training process, denoted as `ControlNet'.
As shown in Tab.~\ref{tab:ablation_adapter} and Fig.~\ref{fig:ablation_adapter}, when the limitation of the adapter on the optimization space is removed, the method is prone to reward hacking during training. 
Although there is a significant improvement in perceptual metrics such as CLIPIQA and MUSIQ, the generated images tend to align with the preferences of these metrics rather than human perception.
This can result in images that are not sufficiently realistic and lack fidelity, such as the strange feathers and the human eye in the sky background in Fig.~\ref{fig:ablation_adapter} (a).
In addition, we employ GPT-4o for comprehensive image quality assessment to determine which model produces images with enhanced naturalness and realism, while utilizing Gram matrices to quantify style discrepancies between generated results and ground truth. 
Compared to the method without the adapter, our method demonstrates superior performance with GPT-4o that prefers our outputs in 84\% of test cases. Furthermore, we achieve a 40.4\% reduction in Gram matrix loss from 0.3051 to 0.1817 relative to GT images.

Similarly, we add the proposed adapter but optimize only the parameters of the adapter during the training process, denoted as `Adapter'.
Our method shows a better trade-off between the reference and non-reference metrics.
When only the adapter is optimized, the parameter space available for optimization is constrained, resulting in a 6.3\%  and 2.0\% decrease in the MANIQA and MUSIQ metrics, respectively.
As can be seen from Fig.~\ref{fig:ablation_adapter}, compared to our method, the model obtained by optimizing only the adapter struggles to generate more refined feathers.

      

\noindent \textbf{Loss Function.}
We begin by evaluating the impact of the chosen loss functions on the results.
As shown in Tab.~\ref{tab:ablation_loss}, when only $\mathcal{L}_{clipiqa}$ is added as the loss function (row 2), the no-reference image quality assessment metrics MANIQA improves by 8.3\% compared to using only the denoising loss (row 1). 
However, reference metrics such as PSNR and SSIM decrease by 4.3\% and 10.6\% respectively, indicating lower fidelity in the generated HR images.
Similarly, when only $\mathcal{L}_{1}$ is added as the loss function (row 3), the reference metrics improve while the non-reference metrics correspondingly decline. 
In contrast, our proposed training strategy, applying $\mathcal{L}_{1}$ and $\mathcal{L}_{clipiqa}$, allows us to enhance image perceptual quality while ensuring structural consistency and fidelity. 
The visualization in Fig.~\ref{fig:ablation_control_weight_map} confirms this as well.
In the initial stages of denoising, the adapter encourages the integration of ControlNet features by increasing the control weight map. 
Subsequently, it progressively reduces these control weights to suppress ControlNet constraints, thereby guiding the generation of high-frequency details to enhance the visual quality.

\begin{table}[!h]
    \centering
    \caption{Ablations of Loss Function.}
    \label{tab:ablation_loss}
    \begin{tabular}{cc cccc} \toprule
      $\mathcal{L}_{1}$
      &$\mathcal{L}_{clipiqa}$
      &$\text{LPIPS}\downarrow$ 
      &$\text{MANIQA}\uparrow$ 
      &$\text{MUSIQ}\uparrow$ 
      &$\text{CLIPIQA}\uparrow$ 
      
  \\ \midrule 
    $\times$   & $\times$     
    &0.3639  
    &0.5781 &67.63   &0.7566  \\
    $\times$   & $\checkmark$     
    &0.4110  
    &0.6265 &69.73   &0.7687  \\
    $\checkmark$    & $\times$      
    &0.3650  
    &0.5586 &67.12   &0.7366  \\ \midrule
    $\checkmark$    & $\checkmark$    
    &0.3762  
    &0.6007 &68.14   &0.7681 \\ \bottomrule
    \end{tabular}
\end{table}

It is noteworthy that when applying $\mathcal{L}_{clipiqa}$, not only does the CLIPIQA metric improve, but all no-reference image quality assessment metrics show an increase. 
During training, both MANIQA and MUSIQ metrics can be employed as image rewards. 
As shown in Tab.~\ref{tab:ablation_reward}, where `MANIQA' denotes that the reward function $\mathbf{R}$ in Equation (\ref{eq:ch3_eq_clipiqa}) corresponds to the MANIQA metrics.
Compared to MUSIQ and MANIQA, CLIPIQA results in a more stable training process and better trade-off. 
Our implementation of CLIPIQA substantiates incorporating the quality metrics, as the image rewards can improve perceptual quality. 
We believe that adopting a better perceptual quality as a reward signal could further enhance the results, but this lies beyond the scope of our research.
\begin{table}[!htb]
    \centering
    \caption{\textbf{Ablations of Timestep-Aware Training Strategy.}}
    \label{tab:ablation_reward}
    \begin{tabular}{c cccc} \toprule
      
      &$\text{LPIPS}\downarrow$ 
      &$\text{MANIQA}\uparrow$ 
      &$\text{MUSIQ}\uparrow$ 
      &$\text{CLIPIQA}\uparrow$ \\ \midrule
      all steps  
    &0.6299  
    &0.5302 &65.12   &0.6790  \\
     inverse steps  
    &0.5830  
    &0.5863 &67.48   &0.7508  \\
    MANIQA  
    &0.4742
    &0.6105 &67.93   &0.7572  \\
    MUSIQ  
    &0.4090  
    &0.6241 &69.23   &0.7651  \\ 	
    CLIPIQA            
    &0.3762  
    &0.6007  &68.14   &0.7681 \\ \bottomrule
    \end{tabular}
\end{table}

\noindent \textbf{Timestep-Aware training Strategy.}
We conducted further experiments to explore the impact of using different reward functions at various time steps.
Specifically, we first added both $\mathcal{L}_{1}$ and $\mathcal{L}_{clipiqa}$ across all time steps (0-1000 steps) in the training process, denoted as "all steps". 
As shown in Tab.~\ref{tab:ablation_reward} and Fig.~\ref{fig:ablation_steps} (a), when image rewards are introduced at all time steps, the model becomes unstable during training and fails to generate the correct HR image. 
Consequently, all reference and non-reference metrics significantly decrease.
Similarly, when we applied the inverse training strategy that introduces $\mathcal{L}_{clipiqa}$ in the early denoising stages (0-800 steps) and $\mathcal{L}_{1}$ in the later stages (0-200 steps),  denoted as "inverse steps", the model also suffered from instability during training, resulting in lower quality high-resolution images.
These experimental results demonstrate that our training strategy is crucial for achieving optimal outcomes.

\section{Conclusion}
\label{sec:cls}
In this paper, we proposed a timestep-aware image super-resolution method that introduces a timestep-aware adapter to integrate ControlNet and diffusion features dynamically. 
In addition, we designed a timestep-aware training strategy to train each module separately based on the generative pattern of the denoising process. 
Extensive experiments on benchmark datasets demonstrate the effectiveness of our method compared with the state-of-the-art methods.

\section*{Acknowledgments}
This research was funded through National Key Research and Development Program of China (Project No. 2022YFB36066), in part by the Shenzhen Science and Technology Project under Grant (KJZD20240903103210014, JCYJ20220818101001004).

\bibliographystyle{ACM-Reference-Format}
\bibliography{acmart}
\end{document}